\documentclass{article}

\usepackage {graphicx}
\usepackage{amsmath} 
\usepackage{hyperref}
\usepackage{amssymb}

\usepackage{geometry}
\usepackage{pdflscape}

\begin{document}
\pagestyle{headings}

\title{Multi Expression Programming for solving classification problems}

\author{Mihai Oltean\\
mihai.oltean@gmail.com, \\
https://mepx.org
}

\maketitle

\begin{abstract}

Multi Expression Programming (MEP) is a Genetic Programming variant which encodes multiple solutions in a single chromosome. This paper introduces and deeply describes several strategies for solving binary and multi-class classification problems within the \textit{multi solutions per chromosome} paradigm of MEP. Extensive experiments on various classification problems are performed. MEP shows similar or better performances than other methods used for comparison (namely Artificial Neural Networks and Linear Genetic Programming).

\end{abstract}

\textbf{Keywords: multi expression programming, multi-class classification, binary classification, genetic programming, data analysis}

\section{Introduction}

Multi Expression Programming (MEP) \cite{mep_tech_report} is an evolutionary technique for automatic generation of computer programs. Specific to MEP is the ability to encode multiple solutions in a single chromosome. All these solutions are analyzed in the same complexity as other techniques encoding a single solution per chromosome. This feature is essential for exploring the search space more efficiently.

So far, MEP implementations were focused on various problems, but the main focus was solving symbolic regression \cite{mep_tech_report} and binary classification problems. A wide range of problems have been attacked so far in this way (see \url{https://mepx.org} for a list of papers on this topic). Specifically, classification problems have been solved with MEP in \cite{Baykasoglu,Telcioglu,Aydogan,Alavi,Zhang,Diosan,Kulluk,Tapkan}.

However, this is not enough, for many real world problems that ask for separating data into more than 2 categories.

Here several new ways to attack classification problems, having two or more classes, are introduced.

These strategies have been already implemented in the \textit{Multi Expression Programming X} software \cite{mepx} in the last years, but no systematic comparison with other techniques has been performed so far. The purpose of this paper is to fill this void.

Paper has the following simple structure: In section \ref{mep}  Multi Expression Programming technique is briefly described. In section \ref{mep_multi_class} several ways to solve multi-class classification problems with MEP are introduced. Methods for strictly binary classification (with 2 classes) (see section \ref{mep_binary}) and, more important, for more than 2 classes (see section \ref{mep_multi_class}) are proposed. Data sets used in the numerical experiments are described in section \ref{test_problems}. Results of the numerical experiments are presented in section \ref{results}. Numerical comparisons with other similar techniques are performed in the same section. At the moment only two comparisons are made: with Linear Genetic Programming and Artificial Neural Networks on multiple problems taken from PROBEN1 \cite{proben1}. Conclusions are drawn in section \ref{conclusions}. Finally we discuss some potential directions for future work.

\section{Multi Expression Programming}
\label{mep}

Multi Expression Programming \cite{mep_tech_report} is a method for  generating computer programs by using ideas inspired from biology. Particularly it can be used for generating programs whose output generates a curve fitting or a classification of data. These problems are complex and have many practical applications including data analysis of data from multiple fields (economy, medicine, engineering etc). 

MEP differentiates from other Genetic Programming \cite{brameier,koza, miller} techniques in several critical points:

\begin{itemize}

\item MEP encodes multiple solutions in the same chromosome. Usually multiple genes are considered as potential candidates for representing the output and the best of them is chosen for the actual fitness assignment. This means that the more solution are investigated and the search space is better explored. Evaluating such chromosome is done (in most of the cases) in the same complexity as in the case of a single solution per chromosome. Some other techniques use the last gene as the output (Linear Genetic Programming \cite{brameier}) and some others evolve the gene providing the output (like Cartesian Genetic Programming \cite{miller}). 

\item MEP evaluates the computer programs instruction by instruction and stores the partial results for all training data. In this way MEP has a huge speed (less \textbf{if}, \textbf{switch} instructions) without needing to work at low level (with processors registers) like Linear Genetic Programming \cite{brameier} does. It is well known that \textbf{if} instruction slows down the program execution \cite{Pikus}.

\end{itemize}

\subsection{MEP representation}

MEP representation is based on \textit{Three-address code} \cite{aho}. However, tree-based representations have been also investigated in the context of multiple solutions per chromosome \cite{mep_linear_representation}.

Each MEP chromosome has a constant number of genes. This number defines the length of the chromosome.
Each gene contains an instruction (also name function) or variable (also named terminal).
A gene that encodes a function includes pointers towards the function arguments. Function arguments always have indexes of lower values than the position of the function itself in the chromosome. This representation ensures that no cycle arises while the chromosome is decoded (phenotypically transcribed). According to the proposed representation scheme, the first symbol of the chromosome must be a terminal symbol. In this way, only syntactically correct programs (MEP individuals) are obtained.
Example

Consider a representation where the numbers on the left positions stand for gene labels. Labels do not belong to the chromosome, as they are provided only for explanation purposes.

For this example we use the set of functions:

$F$ = \{+, *\},

and the set of terminals:

$T$ = \{$a, b, c, d$\}.

An example of chromosome using the sets $F$ and $T$ is given below:

1: $a$

2: $b$

3: + 1, 2

4: $c$

5: $d$

6: + 4, 5

7: * 3, 5

\subsection{Decoding MEP chromosome and the fitness assignment process}

MEP chromosomes are translated into computer programs by parsing them in a top-down fashion. 

\begin{itemize}
\item A terminal symbol specifies a simple expression. 
\item A function symbol specifies a complex expression obtained by connecting the operands specified by the argument positions with the current function symbol.
\end{itemize}

\subsubsection{Example}
For instance, genes 1, 2, 4 and 5 in the previous example encode simple expressions formed by a single terminal symbol. These expressions are:

$E_1 = a$,

$E_2 = b$,

$E_4 = c$,

$E_5 = d$,

Gene 3 indicates the operation + on the operands located at positions 1 and 2 of the chromosome. Therefore gene 3 encodes the expression:

$E_3 = a + b$.

Gene 6 indicates the operation + on the operands located at positions 4 and 5. Therefore gene 6 encodes the expression:

$E_6 = c + d$.

Gene 7 indicates the operation * on the operands located at position 3 and 5. Therefore gene 7 encodes the expression:

$E_7 = (a + b) * d$.

$E_7$ is the last expression encoded by in chromosome.

There is neither practical nor theoretical evidence that one of these expressions is better than the others. This is why each MEP chromosome is allowed to encode a number of expressions equal to the chromosome length (number of genes). The chromosome described above encodes the following expressions:

$E_1 = a$,

$E_2 = b$,

$E_3 = a + b$,

$E_4 = c$,

$E_5 = d$,

$E_6 = c + d$,

$E_7 = (a + b) * d$.

The value of these expressions may be computed by reading the chromosome top down. Partial results are computed by dynamic programming and are stored in a conventional manner.

Due to its multi expression representation, each MEP chromosome may be viewed as a forest of trees rather than as a single tree, which is the case of Genetic Programming.

As MEP chromosome encodes more than one problem solution, it is interesting to see how the fitness is assigned.

\subsubsection{Fitness computation for regression problems}
The chromosome fitness is usually defined as the fitness of the best expression encoded by that chromosome.

For instance, if we want to solve symbolic regression problems, the fitness of each sub-expression $E_i$ may be computed using the formula:

\begin{equation}
\label{eq1}
fitness(E_i ) = \sum\limits_{k = 1}^n {\left| {obtained_{k,i} - target_k } \right|} ,
\end{equation}

where $obtained_k,i$ is the result obtained by the expression $E_i$ for the fitness case $k$ and $target_k$ is the targeted result for the fitness case k. In this case the fitness needs to be minimized.

The fitness of an individual is set to be equal to the lowest fitness of the expressions encoded in the chromosome:

\begin{equation}
\label{eq2}
fitness(C) = \mathop {\min }\limits_i fitness(E_i ).
\end{equation}

When we have to deal with other problems, we compute the fitness of each sub-expression encoded in the MEP chromosome. Thus, the fitness of the entire individual is supplied by the fitness of the best expression encoded in that chromosome.

\subsection{Why encoding multiple solutions within a chromosome?}

When you compute the value of an expression encoded as a GP tree you have the compute the value of all subtrees. This means that all GP subtrees can be viewed as a potential solution of the problem being solved. Most of the GP techniques considers only the one tree while ignoring all the other subtrees. However, the value/fitness for all subtrees is computed by GP. The biggest difference between MEP and other GP techniques is that MEP outputs the best subtree encoded in a chromosome. Note that the complexity (roughly speaking - the running time) is the same for MEP and other GP techniques encoding 1 solution/chromosome.

The second reason for the this question is motivated by the No Free Lunch Theorems for Search \cite{no_free_lunch}. There is neither practical nor theoretical evidence that one of the solutions encoded in a chromosome is better than the others. More than that, Wolpert and McReady proved \cite{no_free_lunch} that we cannot use the search algorithm's behavior so far for a particular test function to predict its future behavior on that function.

\subsection{Why NOT encoding multiple solutions within a chromosome?}

There are also downsides when encoding multiple solutions in a chromosome.
Exploring more points in the search space might lead to overfitting in some cases, which is not good. One can see very good training errors, but bad errors on the test set, so, in such cases, some mechanisms to avoid overfitting are required. Good solutions on the training set does not always translate to good solutions on the test set.

\subsection{Genetic operators}

One cutting point or uniform crossover have been tested with similar results. The cut points are chosen between instructions not inside them.

Some symbols in the chromosome are changed by mutation. To preserve the consistency of the chromosome, its first gene must encode a terminal symbol.

\subsection{Main algorithm - a steady state}

The MEP algorithm starts by creating a random population of individuals.

The following steps are repeated until a stop condition is reached:

\begin{itemize}
    \item Two parents are selected using a standard selection procedure.

\item The parents are recombined in order to obtain two offspring.

\item The offspring are mutated.

\item The fitness of the offspring are computed.

\item Each offspring \textit{O} replaces the worst individual \textit{W} in the current population if \textit{O} is better than \textit{W}.

\end{itemize}

A generation is said to be built when a number individuals equal to the population size is created. 

\section{MEP for binary classification problems}
\label{mep_binary}

For binary classification problems we can use a threshold value for separating the data: everything below threshold is assigned to one class, and everything above is assigned to the other class. The threshold can be either evolved (see section \ref{evolved_threshold}) or automatically computed (see section \ref{automatic_threshold}).

\subsection{Evolved threshold (MEP-BET)}
\label{evolved_threshold}

The threshold is a real value, attached to each chromosome, and is evolved like a any standard real number - by mutation. The threshold is initially randomly generated and then mutated within a range (i.e. we add or subtract a random value within a given range).

The algorithm proceeds as follows: it computes the mathematical value of each MEP formula and then, for each of them, it computes the fitness as follows: for each training data we check if the result of the current expression is below or above the threshold; if it is below, or equal to the threshold, that instance is assigned to class A, otherwise we assign it to B. Of course, the actual class can be different from the computed one and this difference will give us the fitness (which is the number of incorrectly classified data over the number of data).

The short name of this strategy is MEP-BET (MEP for Binary problems with Evolved Threshold).

\subsection{Automatic threshold (MEP-BAT)}
\label{automatic_threshold}

In section \ref{evolved_threshold} we have evolved the threshold. But, this is sometimes a waste of resources, because even if the obtained formula is a good classifier, the threshold could be not optimal, and thus that individual will get a worse fitness.

The question is how we can compute the threshold automatically so that we do not waste resources evolving it.

Here is how: For a given expression in a MEP chromosome we compute its value for each training data. Then, we sort these results (for all training data) in an ascending order. We do that because the order of training data is not important, but the threshold is in that range (between the minimal (MIN) - 1 and maximal (MAX) values of that subexpression).

Now we have the training data sorted by the value of MEP expressions and we must see where exactly the threshold should be inserted. If all data belong to class B, then the threshold should be less than MIN (i.e. MIN - 1). If all data belong to class A, the threshold should be MAX. Otherwise, it is somewhere in between.

We traverse the array of sorted values and we check where the threshold should split the data in order to minimize error. 

We start with a threshold being equal to MIN - 1 meaning that we classify everything to class B. In this case the fitness of that gene is equal to the number of data belonging to class A (that is the number of incorrectly classified data).

We jump to the first (or next) data in the array of sorted values and we assume that the threshold has that value (of the current expression for that input). In this case that training data is labeled with class A (because its value is less or equal to the threshold). If this classification corresponds to the real class, we do nothing. Otherwise we increase the number of incorrectly classified data. Then we move to the next data and so on ... and finally we have the fitness of that gene. The best fitness will be the fitness of the chromosome.

A weak point of this strategy is that is slower than the one which evolves the threshold (described in section \ref{evolved_threshold}). However, the computation of the threshold is exact (for an evolved expression). This is why we will use only this method in the numerical experiments.

The short name of this strategy is MEP-BAT (MEP for Binary problems with Automatic Threshold).

\section{MEP for multi-class classification problems}
\label{mep_multi_class}

The purpose of this section is to describe how multi-class (more than 2 classes) problems can be approached with Multi Expression Programming.
We introduce several fitness assignment procedures for multi-class classification problems: Winner Takes All-Fixed (section \ref{winner_takes_all_multiple_genes_per_output}), Winner Takes All-Smooth (section \ref{smooth_fitness}), Winner Takes All-Dynamic (section \ref{winner_takes_all_best_output}) and Closest Center (section \ref{closest_center}).

\subsection{Winner takes all - fixed (MEP WTA-F)}
\label{winner_takes_all_multiple_genes_per_output}

Recall that the main idea of MEP is to encode multiple solutions in the same chromosome.

Ok, now back to classification.

Imagine that we have to solve a binary classification problem and we have the following MEP chromosome:

$0: a$

$1: b$

$2: + 0, 1$

$3: * 0, 2$

$4: + 1, 2$

$5: / 4, 3$

We employ a classic strategy: \textit{winner takes all} \cite{brameier}. That means that the gene with the highest output value will designate the class to which the current item belongs. 

Usually we have more instructions per chromosome than classes (see previous example where we have 6 instructions and only 2 classes). In such cases we have to select which gene is assigned to (provides the output for) the first class, which gene is assigned for the second class and so on. 

When we work with single solution per chromosome we could use the last \textit{number of classes} genes for our purpose: if the maximal output value is generated by the last instruction it means that the current item belongs to the last class; if the maximal value is generated by the second last, it means that the item belongs to the second last class ... and so on.

But, recall that we work in a framework where a chromosome contains multiple solutions for a problem. In such scenario we use all genes not only the last \textit{number of classes} ones. We can manually assign which genes to be assigned to each class. For instance we can say that genes from even addresses (like 0, 2 and 4) will provide the output for class 0 and genes from odd addresses (like 1, 3 and 5) will provide the output for class \#1, If the maximal output value is generated by genes 0, 2 or 4, then the item is said to belong to class \#0, otherwise it is assigned to class \#1. In such setup we have multiple genes providing the output.

There should be a soft constraint here which enforces the number of genes in a chromosome to be a multiple of the number of classes.

Here is how we do it in more detailed way:

\begin{enumerate}
    \item for a particular example from the training data, we compute the value of all expressions in that chromosome.
    \item  we find the maximal value (out of those 6 values). If there are multiple maximal values, we consider the first occurrence. 
    \item if the maximal value is at an even address (like 0, 2 or 4 in our example), that example from the training data is said to belong to class 0. Otherwise, that data belongs to class \#1.
    \item we now compute the incorrectly classified examples and that is the fitness of the chromosome.
\end{enumerate}

If we have 3 classes, we can say that genes from addresses 0 and 3 will provide the output for class \#0, genes from 1 and 4 will provide output for class \#1 and genes 2 and 5 will provide the output for class \#2.

 Fitness has to be minimized.

\subsection{Winner takes all - Smooth (MEP WTA-S)}
\label{smooth_fitness}

Here we refine the method proposed in section \ref{winner_takes_all_multiple_genes_per_output}.

There is a problem with the method described previously: it is too steep. In many cases there are modifications of the chromosome which does not lead to changes in fitness. We need a smoother way to compute fitness.

First of all we record the minimum and maximum values generated by genes associated with each class. Then we scale all these values to [0, 1] interval. If we do not do this, we can have cases where a very large value overtakes the fitness value.

Let us suppose that the current data row belongs to class \#3. If only the gene(s) associated with class \#3 has (have) the maximal value, then, the fitness is unchanged. That means that the data is correctly classified. 

However, if one or more genes, belonging to other classes, have the same maximal or even bigger values, it means that the current data is not correctly classified. In this case we have to add some penalties depending on how far the maximal value of class \#3 is from the maximal value of other classes (with higher values) and how many other classes have the same or even higher maximal value. More exactly, to the fitness computed in section \ref{winner_takes_all_multiple_genes_per_output} we will add (for each training data) one (1 unit) for each class whose maximal value is at least equal or higher than the maximal value of the current class. Also, we add how far is the maximum of the current class compared to classes whose value is higher (this is the difference between the maximum value of other class and the maximum value of the current one). 

Fitness has to be minimized.

\subsection{Winner takes all - dynamic (MEP WTA-D)}
\label{winner_takes_all_best_output}

In section \ref{winner_takes_all_multiple_genes_per_output} we have described a strategy which uses multiple genes as possible output for a particular class. Moreover, those genes are at fixed position (multiple of class index).

Here we introduce another method which again checks multiple genes for determining the output, but this time, those genes are not preset.

First of all: how we establish that a particular gene classifies the data to a given class? 

We can do this in multiple ways.

For instance, if the output of the gene (for a particular input data) generates a given value (let's say 1), we classify that data as belonging to class \#1 and so on. One weak point here is due to the difficulty to choose a set of numbers which are universally valid for all problems. This is why we have not tested this method.

Another strategy (that we actually use) is to say that a gene $i$ classifies a particular data in class $k$ only if the maximum value (from all genes) has been generated by gene $i$. Thus, during training, we compute (for each data) the maximum value generated by the chromosome's genes. The gene with maximal value is proposed to provide the output for a class. 

Basically we construct a matrix $B$ which has \textit{num genes} rows and \textit{num classes} columns. For each gene $i$ we compute how many data are incorrectly classified if that gene will give the output to class $k$. So, for each data we check if the maximum value is NOT generated by gene $i$ and we add 1 to the value stored on $B(i, k)$ position.

After we constructed the previously described matrix we have to choose \textit{num classes} genes which will provide the desired output (it is obvious that the genes must be distinct unless the outputs are redundant).

In CGP, the genes providing the program's output are evolved just like all other genes. In MEP, the best genes in a chromosome are chosen to provide the program's outputs. When a single value is expected for output we simply choose the best gene (see formulas (\ref{eq1}) and (\ref{eq2})). When multiple genes are required as outputs we have to select those genes which minimize the difference between the obtained result and the expected output. 

\noindent
where $obtained_{k,i}$ is the obtained result by the expression (gene) $E_{i}$ for the fitness case $k$ and $target_{k,q}$ is the targeted result for the fitness case $k$ and for the output $q$. The values $f(E_{i}$, $q)$ are stored in a matrix (by using Dynamic Programming) for latter use (see formula (\ref{eq4})).

Since the fitness needs to be minimized, the quality of a MEP chromosome is computed by using the formula:

\begin{equation}
\label{eq4}
f(C) = \mathop {\min }\limits_{i_1 ,i_2 ,...,i_{NO} } \sum\limits_{k = 1}^{q \le num classes} {B(i ,q)} .
\end{equation}

In equation (\ref{eq4}) we have to choose numbers $i_{1}$, $i_{2}$, \ldots , $i_{num classes}$ in such way to minimize the program's output. For this we shall use a simple heuristic which does not increase the complexity of the MEP decoding process: for each output $q$ (1 $ \le  \quad q \quad  \le $ \textit{NO}) we choose the gene $i$ that minimize the quantity $f(E_{i}$, $q)$. Thus, to an output is assigned the best gene (which has not been assigned before to another output). The selected gene will provide the value of the $q^{th}$ output.

\textbf{\textit{Remark}}:

Formula and (\ref{eq4}) are the generalization of formula (\ref{eq2}) for the case of multiple outputs of a MEP chromosome.

The complexity of the heuristic used for assigning outputs to genes is O(\textit{num genes $ \cdot $ num classes}).

We may use another procedure for selecting the genes that will provide the problem's outputs. This procedure selects, at each step, the minimal value in the matrix $f(E_{i}$, $q)$ and assign the corresponding gene $i$ to its paired output $q$. Again, the genes already used will be excluded from the search. This procedure will be repeated until all outputs have been assigned to a gene. Not that is has a higher complexity -- $O$(\textit{num classes}$ \cdot $\textit{log}$_{2}$(\textit{num classes})$ \cdot $\textit{num genes}) - than the previously described procedure which has the complexity $O$(\textit{num classes}$ \cdot $\textit{num genes}).

\subsection{Closest center - (MEP CC)}
\label{closest_center}

For each class we determine the set of values generated by applying a MEP formula to each data in the training set. Then, we compute the center of mass of the set (for each class; if we have 3 classes we obtain 3 centers). The center of mass was chosen instead of geometric center in order to reduce the influence of the outliers. Then, each data is classified to class with the nearest center. When distances to different centers are equal, the class with the lowest index is chosen. The fitness, of a formula, is the number of incorrectly classified data by that formula.

The computation of centers is made for each instruction (formula) in the chromosome. The best gene (providing the minimum number of incorrectly classified data) gives the fitness of the chromosome.

There is a small added complexity to this approach (multiplication by the number of classes, but this is not significant if you have a small number of classes). But, the most important aspect is that the added complexity does not depend on the chromosome length. So, we can increase the chromosome length without increasing complexity.

\section{Test problems}
\label{test_problems}

Multiple test problems have been used in the numerical experiments performed in this paper. Here we include a description of each dataset along with its characteristics. 

Each dataset is split into 3 parts: Training (50\%), Validation (25\%) and Test (25\%).

For the problems taken from PROBEN1 \cite{proben1}, three variations of the same dataset are generated by data permutation.

\textbf{Cancer - Diagnosis of breast cancer}

"\textit{Try to classify a tumor as either benign or malignant based on cell descriptions gathered by microscopic examination. Input attributes are for instance the clump thickness,
the uniformity of cell size and cell shape, the amount of marginal adhesion, and the frequency of bare nuclei.}" Data are taken from PROBEN1 dataset \cite{proben1}.

\textbf{Card - Predict the approval or non-approval of a credit card to a customer}

"\textit{Each example represents a real credit card application and the output describes whether the bank (or similar institution) granted the credit card or not.}" Data are taken from PROBEN1 dataset \cite{proben1}.

\textbf{Diabetes - Diagnose diabetes of Pima indians}

"\textit{Based on personal data (age, number of times pregnant) and the results of medical examinations (e.g. blood pressure, body mass index, result of glucose tolerance test, etc.), try to decide whether a Pima indian individual is diabetes positive or not.}" Data are taken from PROBEN1 dataset \cite{proben1}.

\textbf{Gene - Detect intron/exon boundaries (splice junctions) in nucleotide sequences} 

"\textit{From a window of 60 DNA sequence elements (nucleotides) decide whether the middle is either an intron/exon boundary (a donor), or an exon/intron boundary (an acceptor), or none of these.}" Data are taken from PROBEN1 dataset \cite{proben1}.

\textbf{Glass - Classify glass types}

"\textit{The results of a chemical analysis of glass splinters (percent content of 8 different elements) plus the refractive index are used to classify the sample to be either float processed or non
float processed building windows, vehicle windows, containers, tableware, or head lamps.}" Data are taken from PROBEN1 dataset \cite{proben1}.

\textbf{Heart - Predict heart disease}

"\textit{Decide whether at least one of four major vessels is reduced in diameter by more than 50\%. The binary decision is made based on personal data such as age, sex, smoking habits, subjective patient pain descriptions, and results of various medical examinations such as blood pressure and electro cardiogram results.}" Data are taken from PROBEN1 dataset \cite{proben1}.

\textbf{Heartc - Predict heart disease}

It is subset of \textbf{Heart} dataset (see above) having the cleanest data (only few values are missing). Data are taken from PROBEN1 dataset \cite{proben1}.

\textbf{Horse - Predict the fate of a horse that has a colic}

"\textit{The results of a veterinary examination of a horse having colic are used to predict whether the horse will survive, will die, or will be euthanized.}" Data are taken from PROBEN1 dataset \cite{proben1}.

\textbf{Soybean - Recognize 19 different diseases of soybeans}

"\textit{The discrimination is done based on a description of
the bean and the plant plus
information about the history of the plant's life.}" Data are taken from PROBEN1 dataset \cite{proben1}.

\textbf{Thyroid - Diagnose thyroid hyper- or hypofunction}

"\textit{Based on patient query data and patient examination data, the task is to decide whether the patient's thyroid has overfunction, normal function, or underfunction.}" Data are taken from PROBEN1 dataset \cite{proben1}.

Characteristics of each data set are summarized in Table~\ref{tab:dataset_summary}.

\begin{table}[htbp]
\caption{Summarization of dataset used in numerical experiments.}
\label{tab:dataset_summary}
\begin{tabular}{ l l l l l }
  Problem & Num attributes & Num classes & Num examples & (Training + Validation + Test)\\
  \hline
  Cancer & 9 & 2 & 699 & 350+175+174\\
  Card & 51 & 2 & 690 & 345+173+172\\
  Diabetes & 8 & 2 &690 & 384+192+192\\
  Gene & 120 & 3 & 3175 & 1588+794+793\\
  Glass & 9 & 6 & 214 & 107+54+53\\
  Heart & 35 & 2 & 920 & 460+230+230\\
  Heartc & 35 & 2 & 303 & 152+76+75\\
  Horse & 58 & 3 & 364 & 182+91+91\\
  Soybean & 82 & 19 & 683 &342+171+170\\
  Thyroid & 21 & 3 &7200 &3600+1800+1800\\
  \hline
\end{tabular}
\end{table}

\section{Numerical experiments and comparison with other techniques}
\label{results}

Various experiments were performed with MEP on different problems.  Main values for parameters were similar to those reported in other papers used for comparison. For instance, in \cite{brameier} Linear Genetic Programming is compared with Artificial Neural Networks and, in this case, most of the parameters (but not all) were copied from the LGP. The reader is invited to study the details about MEP parameters in each section where MEP is compared to another technique.

Note that is difficult to make a fair comparison with other techniques due to multiple reasons:

\begin{itemize}

\item we did not have access to the source code for the other techniques, so we directly compared only with the published results.

\item some parameters of other methods have no correspondent within MEP.

\item data shuffling and splitting is not perfectly known in all cases.

\end{itemize}

MEP settings are given in Table \ref{tab:MEP_settings}.

\begin{table}[htbp]
\caption{MEP parameters for MEP vs LGP and MEP vs ANNs experiments (see \cite{brameier} for more details).}
\label{tab:MEP_settings}
\begin{tabular}{ l l }
  Parameter & Value\\
  \hline
  Sub population size & 500\\
  Number of sub populations & 10\\
  Sub populations architecture & ring\\
  Migration rate & 1 (per generation)\\ 
  Chromosome length&256\\
  Crossover probability&0.9\\
  Mutation probability&0.005\\
  Tournament size&2\\
  Functions probability&0.5\\
  Variables probability&0.4\\
  Constants probability&0.01\\
  Number of generations&250\\
  Mathematical functions&+,-,*, /, sin, exp, ln, a$<$b?c:d\\
  Number of constants&10\\
  Constants initial interval&randomly generated over [0, 1]\\
  Constants can evolve?&YES\\
  Constants can evolve outside initial interval?&YES\\
  Constants delta& 0.1\\

  \hline
\end{tabular}
\end{table}

Results of the experiments are given in Table \ref{tab:results_mep_classification}. In our tests we have not included the algorithm which evolves the threshold (described in section \ref{evolved_threshold}) because it was generally worse than the one which computes the threshold automatically. Still, please note that evolving the threshold is faster than computing it automatically, so, in some cases, it still make sense to use that strategy in some particular cases (where time matters).

\begin{landscape}
\begin{table}[htbp]
\caption{Results on various problems. 30 runs are performed. Number of incorrectly classified data is given in percent. \textit{Best} stands for the best solution (out of 30 runs), \textit{Avg} stands for \textit{Average} (over 30 runs) and \textit{Dev} stands for \textit{Standard Deviation}. Best results are bolded. N/A stands for Not Apply and means that the corresponding algorithm cannot be applied (the algorithm is for 2 classes and the problem has more than 2 classes). Values have been truncated after the $2^{nd}$ decimal.}
\label{tab:results_mep_classification}
\begin{tabular}{|l|lll|lll|lll|lll|lll|}
  \hline
Problem& \multicolumn{3}{c}{MEP WTA-F}& \multicolumn{3}{|c|}{MEP WTA-S}	&\multicolumn{3}{c}{MEP WTA-D} &\multicolumn{3}{|c|}{MEP CC}&\multicolumn{3}{|c|}{MEP BAT}\\
  \hline

&Best&Avg&Dev&Best&Avg&Dev&Best&Avg&Dev&Best&Avg&Dev&Best&Avg&Dev	\\
  \hline
  \hline
cancer1&1.14&2.62&1.03&1.14&2.79&0.86&0.57&\textbf{2.2}&0.81&0.57&2.33&0.71&\textbf{0}&2.33&0.78	\\
\hline
cancer2&4.02&6.14&1.41&4.02&6.28&1.25&4.02&5.97&0.82&5.17&6.14&0.36 &\textbf{3.44}&\textbf{5.7}&0.9 \\
\hline
cancer3&2.87&4.98&1.25&2.87&\textbf{4.9}&1.13&3.44&5&1.09&\textbf{2.44}&5.17&0.95 &3.44&5.09&1 \\
\hline
\hline
card1&\textbf{11.62}&14.84&1.38&12.2&14.57&0.96&12.79&14.34&0.92&11.62&14.55&0.81&12.2&\textbf{13.79}&1.09 \\
\hline
card2&17.44&20.17&1.41&18.02&19.88&1.23&\textbf{12.79}&\textbf{14.34}&0.92&18.60&20.34&1.35&18.02&20.94&1.29\\
\hline
card3&15.69&17.94&1.05&\textbf{12.79}&17.69&1.49&13.37&\textbf{16.99}&1.24&15.69&17.86&0.71&15.11&17.48&1.44 \\  
\hline
\hline
diabetes1&20.83&24.35&2.02&20.83&25.43&1.59&\textbf{20.31}&24.44&2.01&21.35&\textbf{23.54}&1.49&21.35&25.43&1.98\\
\hline
diabetes2&25.52&28.97&1.65&25&29.28&1.95&24.47&27.34&1.71&\textbf{23.43}&\textbf{26.05}&1.51&23.43&26.9&1.9\\
\hline
diabetes3&\textbf{19.79}&24.46&1.94&21.35&24.42&2.04&20.83&23.38&1.26&20.31&\textbf{22.58}&1.59&20.31&23.62&1.43 \\  
\hline
\hline
gene1&\textbf{8.44}&\textbf{10.64}&1.76&7.94&12.02&2.24&9.20&11.15&1.52&10.34&13.22&2.54 &N/A&N/A&N/A \\  
\hline
gene2&8.19&\textbf{9.78}&1.4&7.18&10.5&1.62&7.81&10.2&1.66&\textbf{6.17}&10.61&2.2 &N/A&N/A&N/A \\  
\hline
gene3&10.08&\textbf{12.31}&1.47&\textbf{9.2}&12.94&1.97&10.71&13.38&1.81&10.21&14.16&3.15&N/A&N/A&N/A \\  
\hline
\hline
glass1&\textbf{26.41}&\textbf{36.47}&4.78&28.3&37.92&8.09&32.07&42.2&6.17&30.18&38.93& 5.14&N/A&N/A&N/A \\  
\hline
glass2&30.18&36.66&3.1&30.18&\textbf{35.84}&3&30.18&37.98&5.19&\textbf{28.3}&41&5.22 &N/A&N/A&N/A \\  
\hline
glass3&30.18&38.42&5.3&\textbf{28.3}&\textbf{36.41}&6.8&28.3&49.37&13.82&30.18&37.79&4.95 &N/A&N/A&N/A \\  
\hline
\hline
heart1&\textbf{15.21}&\textbf{21.5}&3.02&16.95&22.62&2.7&20&22.95&1.4&20&22.42&1.56&18.69&21.86&1.48 \\  
\hline
heart2&18.69&21.68&2.02&\textbf{17.39}&21.95&2.3&18.26&\textbf{21.63}&1.46&18.69&22.31&1.41&19.56&22.56&1.59\\  
\hline
heart3&\textbf{22.17}&\textbf{26.31}&1.78&23.04&27.68&2.27&23.91&26.62&1.28&23.91&26.85&1.61&23.91&26.79&1.81 \\
\hline
\hline
heartc1&20&23.24&1.96&18.66&24.17&2.52&17.33&20.13&1.47&17.33&19.91&0.83&\textbf{16}&\textbf{19.77}&1.14 \\  
\hline
heartc2&2.66&7.73&3.21&1.33&9.11&3.8&\textbf{1.33}&\textbf{3.95}&2.76&1.33&4.71&3.2&1.33&6.71&2.67\\  
\hline
heartc3&13.33&17.11&1.61&12&16.75&2.05&\textbf{10.66}&\textbf{13.68}&1.85&12&13.82&1.8&12&14.48&2.08 \\    
\hline
\hline
horse1&26.37&31.94&2.85&\textbf{24.17}&\textbf{29.78}&3.30&28.57&32.08&2.60&28.57&31.72&2.09&N/A&N/A&N/A \\  
\hline
horse2&32.96&37.47&3.54&31.86&37.36&2.73&\textbf{29.67}&36.77&2.31&31.86&\textbf{35.53}&1.93&N/A&N/A&N/A \\  
\hline
horse3&31.86&36.44&3.19&29.67&36.37&3.5&31.86&35.71&2.38&\textbf{28.57}&\textbf{34.79}&2.82&N/A&N/A&N/A \\  
\hline
\hline
soybean1&9.41&17.9&3.89&15.88&30.84&6.55&\textbf{8.82}&\textbf{12.64}&2.17 &15.88&23.27&3.72&N/A&N/A&N/A \\  
\hline
soybean2&11.76&19.82&4.46&17.05&29.7&6.85&\textbf{5.88}&\textbf{10.96}&3.07&11.17&19.45&3.66&N/A&N/A&N/A \\  
\hline
soybean3&12.35&19.01&4.41&14.70&31.5&8.59 &\textbf{7.64}&\textbf{10.23}&3.38&14.11&26&4.33 &N/A&N/A&N/A \\  
\hline
\hline
thyroid1&\textbf{0.83}&\textbf{1.46}&0.29&0.88&2.03&0.64&1.22&1.8&0.36&1.44&2.04&0.29 &N/A&N/A&N/A \\  
\hline
thyroid2&0.61&\textbf{0.98}&0.34&\textbf{0.55}&1.12&0.56&0.61&1.05&0.28&0.83&1.58&0.25&N/A&N/A&N/A \\  
\hline
thyroid3&0.5&1.3&0.49&\textbf{0.5}&\textbf{1.09}&0.41&0.5&1.24&0.42&1.27&1.66&0.19 &N/A&N/A&N/A \\  
\hline
\end{tabular}
\end{table}
\end{landscape}

Running times are given in Table \ref{tab:runtime_mep_classification}. Only the best running time was shown because the computer where the experiments were performed had to be put in standby mode at various times, thus the longest running time is not accurate. However, very small variations in running times between different runs have been noticed, so presenting only the fastest one does not change the conclusions. Also, the running time are not so important in the near future because processor will improve. What is more important is to see the magnitude and the differences between different MEP strategies.

\begin{table}[htbp]
\caption{Runtime (in seconds) on various problems. 30 runs are performed and the best time is shown in this table. Fasterst algorithms are bolded. N/A stands for Not Apply and means that the corresponding algorithm cannot be applied (the algorithm is for 2 classes and the problem has more than 2 classes). Values have been truncated after the $2^{nd}$ decimal.}
\label{tab:runtime_mep_classification}
\begin{tabular}{|l|l|l|l|l|l|}
  \hline
Problem& MEP WTA-F& MEP WTA-S& MEP WTA-D &MEP CC&MEP BAT\\
  \hline

  \hline
cancer1&\textbf{96.25}&131.92&97.86&297.74&383.38\\
\hline
\hline
card1&\textbf{89.76}&119.62&96.86&325.07&186.63\\
\hline
\hline
diabetes1&\textbf{35.3}&45.19&93.37&125.7&268.95\\
\hline
\hline
gene1&\textbf{301.89}&624.25&486.13&2079.56&N/A \\  
\hline
glass1&\textbf{29.88}&44.71&39.49&157.88&N/A \\  
\hline
heart1&\textbf{114.16}&157.82&115.64&379.42&311.17\\  
\hline
heartc1&\textbf{36.71}&58.47&39.99&106.52&93.32\\  
\hline
horse1&\textbf{53.38}&74.97&57.3&207.78&N/A \\  
\hline
soybean1&\textbf{84.89}&121.19&112.36&937.74&N/A \\  
\hline
thyroid1&\textbf{776.42}&1197.2&842.36&2252.1&N/A \\  
\hline
\end{tabular}
\end{table}

One can see that MEP WTA-F is the fastest algorithm and MEP-CC and MEP-BAT are the slowest one. For the \textbf{Soybean} problem (which has 19 classes), the MEP-CC is one order of magnitude slower than the MEP WTA-F. This clearly shows the influence of the number of classes in the complexity of the MEP CC algorithm.

\subsection{Comparison with Linear Genetic Programming}
\label{mep_vs_lgp}

In \cite{brameier} a comparison between Linear Genetic Programming (LGP) and Artificial Neural Networks is performed. In this section we describe a similar comparison between Linear Genetic Programming and Multi Expression Programming. Problems are taken from \cite{proben1}.

For a fair comparison, we tried to use similar settings for MEP as those given in \cite{brameier} for LGP. But we could not match them perfectly due to multiple reasons, which are detailed below:

\begin{itemize}
    \item 
The main parameter that we completely disagree is the mutation probability. The authors of paper \cite{brameier} have used a mutation probability of 0.9 (90\% - see Table IV from their paper). That means that the mutation changes 90\% of the chromosome. That means that code sequences inherited from parents are (almost) completely changed after each mutation and thus the purpose of the crossover is significantly diminished. In our experiments, performed with MEP, we used a low mutation probability (less than 1\%). We tried to used higher mutation probabilities but we have obtained worse results. Also, too low mutation probabilities are bad for MEP (due to the lack of novelty), so the only solution is to find the sweet value with experimentation. Definitively, mutation probability is a parameter that deserves full attention.
    \item 
Another setting that we did not agree and we did not copy it, is related to the constants. LGP \cite{brameier} uses integer numbers from 0 to 255 (256 constants). We did not understand why this set of numbers was used. The inputs of most of the problems were real numbers between 0 and 1 or Boolean values. In such setup it is hard to understand what is the purpose of the constant 255. In our experiments with MEP we have used real-valued constants randomly generated over the [0, 1] interval. Constants can evolve and may get values outside of the initial interval, so if needed, the program can evolve higher-value constants (like 255 in LGP setup).
    \item 
Another aspect related to the constants is how many to use. In our opinion, Genetic Programming should use less constants, because the purpose of GP is to generate human-readable mathematical expressions. This is the main advantage of GP over ANNs: readability. How readable is a formula full of constants? ANNs are made of constants and because of that, are hard to interpret. We do not want to obtain hard to interpret formulas. This is why we have limited the number of constants in our experiments to 10. We did not try with another number of constants due to the high time required for a run to complete in the case of problems with many training data.
    \item 
An LGP feature does not have a corresponding feature in MEP: the skip instruction. This LGP specific instruction skips the next instruction if some condition is met. Instead, in MEP experiments, another form of the \textit{if} instruction: $if a < b\ result = c;\ else\ result = d;$ has been added.

\end{itemize}

The results of the comparison are given in Table \ref{tab:comparison_MEP_LGP}. LGP results are taken from \cite{brameier}. For MEP we have taken the best result presented in section \ref{tab:results_mep_classification}. We took the best result generated by one of the MEP strategies (4 or 5) instead of one of them, because anyway we can create an assemble of strategies which outputs the best solution. The time consuming part of MEP is chromosome evaluation on all data, not the class-computation part, so creating an assemble will not increase complexity (compared to the highest-complexity strategy so far).

Note that problem \textbf{heart} from \cite{brameier} is actually the problem \textbf{heartc} from \cite{proben1}. This issue has been identified by looking to the number of training data specified in the mentioned papers.

A difference Delta ($\Delta$) between MEP and LGP is computed as indicated in \cite{brameier}. The actual formula is:

\begin{equation}
\Delta = \frac{(LGP - MEP)}{max\{MEP, LGP\}} * 100
\label{eq_mep_Vs_lgp}    
\end{equation}

Negative values for delta means that MEP is perform worse than LPG. Positive values means that MEP is better. The result is always inside the [-100, 100] interval.

\begin{table}[htbp]
\caption{MEP results compared to LGP. \textit{MEP best} represents the best result obtained by one of the 4 (or 5 in the case of binary classification) MEP methods. MEP data are taken from Table \ref{tab:results_mep_classification}. Best results are bolded.}
\label{tab:comparison_MEP_LGP}
\begin{tabular}{|l| l l| l l|l|}
\hline

Problem&\multicolumn{2}{|c|}{LGP}&\multicolumn{2}{|c|}{MEP best}&$\Delta$ (\%)| \\
\hline
 &Average&StdDev&Average&StdDev&\\
\hline
\hline
cancer1&\textbf{2.18}&0.59&2.2&0.81&-0.9 \\
\hline
cancer2&5.72&0.66&\textbf{5.7}&0.9&+0.34\\
\hline
cancer3&4.93&0.65&\textbf{4.9}&1.13&+0.6\\
\hline
\hline
diabetes1&23.96&1.42&\textbf{23.54}&1.49&+1.75\\
\hline
diabetes2&27.85&1.49&\textbf{26.05}&1.51&+6.46\\
\hline
diabetes3&23.09&1.27&\textbf{22.58}&1.59&+2.2\\
\hline
\hline
gene1&12.97&2.24&\textbf{10.64}&1.76&+17.96\\
\hline
gene2&11.95&2.15&\textbf{9.78}&1.4&+18.15\\
\hline
gene3&13.84&2.09&\textbf{12.31}&1.47&+11.05\\
\hline
\hline
heartc1&21.12&2.02&\textbf{19.7}&1.14&+6.72\\
\hline
heartc2&7.31&3.31&\textbf{3.95}&2.76&+45.96\\
\hline
heartc3&13.98&2.03&\textbf{13.68}&1.85&+2.14\\
\hline
\hline
horse1&30.55&2.24&\textbf{29.78}&3.3&+2.52\\
\hline
horse2&36.12&1.95&\textbf{35.53}&1.93&+1.63\\
\hline
horse3&35.44&1.77&\textbf{34.79}&2.82&+1.83\\
\hline
\hline
thyroid1&1.91&0.42&\textbf{1.46}&0.29&+23.56\\
\hline
thyroid2&2.31&0.39&\textbf{0.98}&0.34&+57.57\\
\hline
thyroid3&1.88&0.36&\textbf{1.09}&0.41&+42.02\\
\hline
\hline
\end{tabular}
\end{table}

One can see that MEP performs better than LGP on all problems except \textbf{cancer1}. Differences are bigger on problems with many training data (gene and thyroid). This could be due to lower overfitting of MEP (Recall that a strategy to reduce overfitting is to use a big number of training data).

\subsection{Comparison with Artificial Neural Networks}

Results for ANNs are taken from \cite{proben1}. Note that the ANNs from \cite{proben1} have been adapted to each problem whereas MEP settings are general for all problems. The results of the comparison are given in Table \ref{tab:comparison_MEP_ANNs}. A difference Delta ($\Delta$) - as described in section \ref{mep_vs_lgp}, equation \ref{eq_mep_Vs_lgp} is computed.

\begin{table}[htbp]
\caption{MEP results compared to ANNs. \textit{MEP best} represents the best/worst result obtained by one of the 4 (or 5) MEP methods. MEP data are taken from Table \ref{tab:results_mep_classification}. Best results are bolded.}
\label{tab:comparison_MEP_ANNs}
\begin{tabular}{|l| l l| l l|l|}
\hline

Problem& \multicolumn{2}{|c|}{ANNs} &\multicolumn{2}{|c|}{MEP best} &$\Delta$ (\%)\\
\hline
 &Average&StdDev&Average&StdDev&\\
\hline
cancer1&\textbf{1.38}&0.49&2.2&0.81&-37.27 \\
\hline
cancer2&\textbf{4.77}&0.94&5.7&0.9&-16.31\\
\hline
cancer3&\textbf{3.70}&0.52&4.9&1.3&-24.48\\
\hline
card1&14.05&1.03&\textbf{13.79}&1.09&+1.85\\
\hline
card2&18.91&0.86&\textbf{14.34}&0.92&+24.16\\
\hline
card3&18.84&1.19&\textbf{16.99}&1.24&+9.81\\
\hline
diabetes1&24.10&1.91&\textbf{23.54}&1.49&+2.32\\
\hline
diabetes2&26.42&2.26&\textbf{26.05}&1.51&+1.4\\
\hline
diabetes3&22.59&2.23&\textbf{22.58}&1.59&+0.04\\
\hline
gene1&16.67&3.75&\textbf{10.64}&1.76&+36.17\\
\hline
gene2&18.41&6.93&\textbf{9.78}&1.4&+46.87\\
\hline
gene3&21.82&7.53&\textbf{12.31}&1.47&+45.58\\
\hline
glass1&\textbf{32.70}&5.34&36.47&4.78&-10.33\\
\hline
glass2&55.57&3.70&\textbf{35.84}&3&+35.5\\
\hline
glass3&58.40&7.82&\textbf{36.41}&6.8&+37.65\\
\hline
heart1&\textbf{19.72}&0.96&21.5&3.02&-8.27\\
\hline
heart2&\textbf{17.52}&1.14&21.63&1.46&-19\\
\hline
heart3&\textbf{24.08}&1.12&26.31&1.78&-8.47\\
\hline
heartc1&20.82&1.47&\textbf{19.7}&1.14&+5.37\\
\hline
heartc2&5.13&1.63&\textbf{3.95}&2.76&+23\\
\hline
heartc3&15.40&3.20&\textbf{13.68}&1.85&+11.16\\
\hline
horse1&\textbf{29.19}&2.62&29.78&3.3&-1.98\\
\hline
horse2&35.86&2.46&\textbf{35.53}&1.93&+0.92\\
\hline
horse3&\textbf{34.16}&2.32&34.79&2.82&-1.81\\
\hline
soybean1&29.40&2.50&\textbf{12.64}&2.17&+57\\
\hline
soybean2&\textbf{5.14}&1.05&10.96&3.07&-53.1\\
\hline
soybean3&11.54&2.32&\textbf{10.23}&3.38&+11.35\\
\hline
thyroid1&2.38&0.35&\textbf{1.46}&0.29&+38.65\\
\hline
thyroid2&1.91&0.94&\textbf{0.98}&0.34&+48.69\\
\hline
thyroid3&2.27&0.32&\textbf{1.09}&0.41&+51.98\\
\hline
\end{tabular}
\end{table}

One can see that MEP performs better than ANNs on more problems than viceversa (20 vs 10). Soybean problem are interested to follow in order to find why there are such change in results between different permutations of data.

\section{Conclusions and future work}
\label{conclusions}

Several strategies for binary and multi-class classification problems have been proposed for the Multi Expression Programming. All strategies follow the basic MEP idea of encoding multiple solutions in a chromosome and choosing the best one for the fitness assignment purposes.

Multiple numerical experiments have been performed on classical problems taken from popular benchmarks. MEP has been compared to other GP techniques and with Artificial Neural Networks.
Numerical experiments shows that MEP can perform well and sometimes even better than the compared techniques.

The idea of encoding multiple solutions in a chromosome is appealing, but it has its own drawbacks: more efficient exploration of the training set can lead to overfitting, which in turn lead to poorer generalization capabilities on new data (the test set). This is why, future work should be focused on just that: how to create a balance between efficient exploration and generalization.

Future work will be focused on implementing more strategies for classification in MEP. Some strategies proposed in conjunction with other GP techniques can be adapted to multi-solution per chromosome strategy of MEP, thus they can constitute a good starting point.

Another development direction will be focused on combining multiple strategies into a single one. The most time consuming function in MEP is, by far, the one computing the value of each gene for each training data. All other functions for computing the class can add only a constant extra computational time, thus, combining multiple techniques (in the sense that more are applied and the one with minimal validation error is chosen) could be an appealing approach.

\section*{Acknowledgments}

All experiments were performed using MEPX software which can be freely downloaded from either \url{https://www.mepx.org}, \url{https://mepx.github.io} or \url{https://github.com/mepx}. Source code is also available on the same site.

All test projects are also available for download.

\end{document}